\pdfoutput=1

\documentclass[11pt]{article}

\usepackage{acl}

\usepackage{times}
\usepackage{latexsym}

\usepackage[T1]{fontenc}

\usepackage[utf8]{inputenc}

\usepackage{microtype}
\usepackage{times}
\usepackage{latexsym}
\usepackage{stmaryrd}
\usepackage{amsmath}
\usepackage{multirow}
\usepackage{bbm}
\usepackage{graphicx}
\usepackage{amssymb}
\usepackage{booktabs}
\usepackage{adjustbox}
\usepackage{xspace}
\usepackage{algpseudocode}
\usepackage{algorithm}
\usepackage{graphicx}
\usepackage{caption}
\usepackage{subcaption}
\usepackage{color, colortbl}
\usepackage{tablefootnote}
\usepackage{enumitem}  
\usepackage{lipsum}
\usepackage{xspace}

\definecolor{mygreen}{RGB}{16, 128, 48}

\setlist{topsep=3pt,itemsep=2pt,partopsep=2pt, parsep=2pt}

\title{InstructAlign: High-and-Low Resource Language Alignment \\via Continual Crosslingual Instruction Tuning}

\author{
  Samuel Cahyawijaya, Holy Lovenia, Tiezheng Yu, Willy Chung, Pascale Fung \\
  Hong Kong University of Science and Technology \\
  Clear Water Bay, Hong Kong \\
  \texttt{scahyawijaya@connect.ust.hk} \\ 
} 

\begin{document}
\maketitle

\begin{abstract}


Large language models (LLMs) that are tuned with instructions have demonstrated remarkable capabilities in various tasks and languages. However, their ability to generalize to underrepresented languages is limited due to the scarcity of available data. Additionally, directly adapting new languages to instruction-tuned LLMs can result in catastrophic forgetting, which leads to the loss of multitasking ability. To address this issue, we propose InstructAlign which uses continual crosslingual instruction tuning to enable LLMs to align new unseen languages with previously learned high-resource languages. Our results demonstrate the effectiveness of InstructAlign in enabling the model to understand low-resource languages with limited parallel data while preventing catastrophic forgetting. Our work contributes to the advancement of language adaptation methods, particularly for adapting instruction-tuned LLMs to underrepresented languages. Our code is released on \url{https://github.com/HLTCHKUST/InstructAlign}.

\end{abstract}

\section{Introduction}

\begin{figure*}[!t]
    \centering
    \resizebox{0.95\linewidth}{!}{
    \includegraphics[clip,trim={0 1em 0 0}]{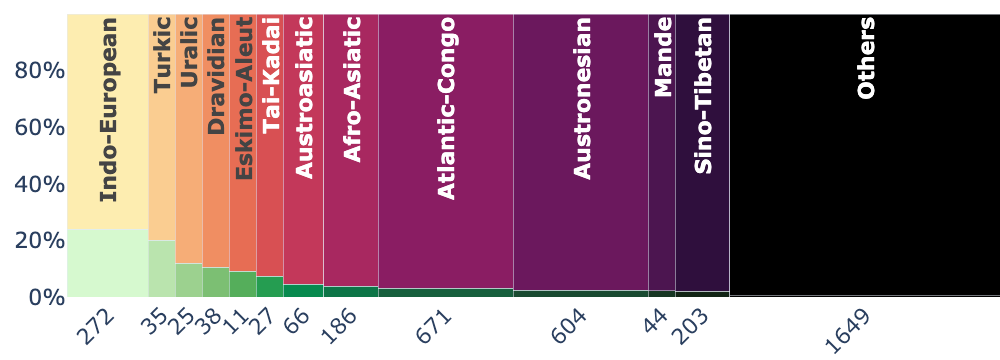}
    }
    \caption{
    The number of languages supported by existing LLMs (\textcolor{mygreen}{green region}) per language family\footnotemark. Existing LLMs only support a fraction of languages around the globe. Most of them are within the Indo-European language family, while most other language families are underrepresented or even unexplored. 
    }
    \label{fig:intro}
\end{figure*}

Instruction-tuned Large language models (LLMs) have demonstrated their generalization capability of solving various tasks expressed in natural language without requiring any explicit training on the corresponding task~\cite{brown2020gpt3,smith2022megatron,rae2022scaling,thoppilan2022lamda,chowdhery2022palm,scao2022bloom,zeng2022glm130b}. This generalization capability is further improved with various tuning methods, such as instruction tuning~\cite{sanh2022t0,wei2022flan,chung2022flan-t5,muennighoff2022crosslingual}. 
However, LLMs and their instruction-tuned variants face difficulties in generalizing across various languages, leading to a disparity in performances\cite{xue2021mt5,gehrmann-etal-2022-gemv2,
scao2022bloom,chowdhery2022palm,yong2023prompting,zhang2023multilingual,asai2023buffet,kabra-etal-2023-multi}. Moreover, these models have limited language coverage, mostly in the Indo-European language family as indicated in Figure~\ref{fig:intro}. For instance, BLOOM~\cite{scao2022bloom} and BLOOMZ~\cite{muennighoff2022crosslingual}, the largest community-driven open-source multilingual pre-trained LLM, only covers 46 languages during pre-training, excluding some high-resource languages with hundreds of millions of speakers, such as German, Japanese, Korean, and Russian, as well as many more low-resource languages with millions of speakers, such as Serbian, Finnish, Amharic, Sinhala, Lao, Javanese, Sundanese, etc.

Expanding the language repertoire of LLMs is essential for promoting inclusivity and diversity in Natural Language Processing (NLP) technology, particularly for languages that are underrepresented and low-resource. Recent studies, including ~\citet{wilie2020indonlu,cahyawijaya2021indonlg,aji2022one,adelani2021masakhaner,adelani2022masakhaner2,kakwani2020indicnlpsuite,kumar2022indicnlg,ebrahimi2022americasnli,adilazuarda-etal-2022d-indorobusta,cahyawijaya2023nusacrowd,cahyawijaya2023nusawrites,song2023globalbench} have emphasized the importance of this issue. 
To address this concern, previous research ~\cite{yong2022bloom1}  has demonstrated that continual pretraining~\cite{chau2020parsing,muller2021unseen,ebrahimi2021adapt} and parameter-efficient fine-tuning (PEFT) methods, like MAD-X~\cite{pfeiffer2020mad} and (IA)$^3$~\cite{liu2022fewshot}, can be utilized to swiftly integrate the knowledge of unseen languages into LLMs using monolingual corpora of the new languages by conducting masked language modeling (MLM)~\cite{devlin2019bert}. However, these methods become ineffective when applied directly to instruction-tuned LLMs due to catastrophic forgetting~\cite{robert1993catastrophic} which prevents them from solving general natural language tasks after the language adaptation phase~\cite{yong2022bloom1}. Moreover, adapter-based approaches, such as MAD-X~\cite{pfeiffer2020mad}, result in the loss of multilingual inference capability due to modularity~\cite{adilazuarda2023the}.


To solve this problem, we introduce InstructAlign, a continual instruction tuning framework to seamlessly align newly adapted low-resource languages (\textcolor{teal}{L2}) with the pre-trained high-resource languages (\textcolor{red}{L1}) of an instruction-tuned LLM through crosslingual alignment. InstructAlign compels LLMs to perform crosslingual alignments between pre-trained and novel languages through alignment-based crosslingual instruction tuning, enabling the model to grasp \textcolor{teal}{L2} with only a limited amount of parallel data. To further prevent catastrophic forgetting, InstructAlign incorporates experience replay~\cite{chaudhry2019tiny,david2019experiencereplay}, which adds past data during the instruction tuning.

\begin{figure*}[!t]
    \centering
    \resizebox{\linewidth}{!}{
    \includegraphics[clip]{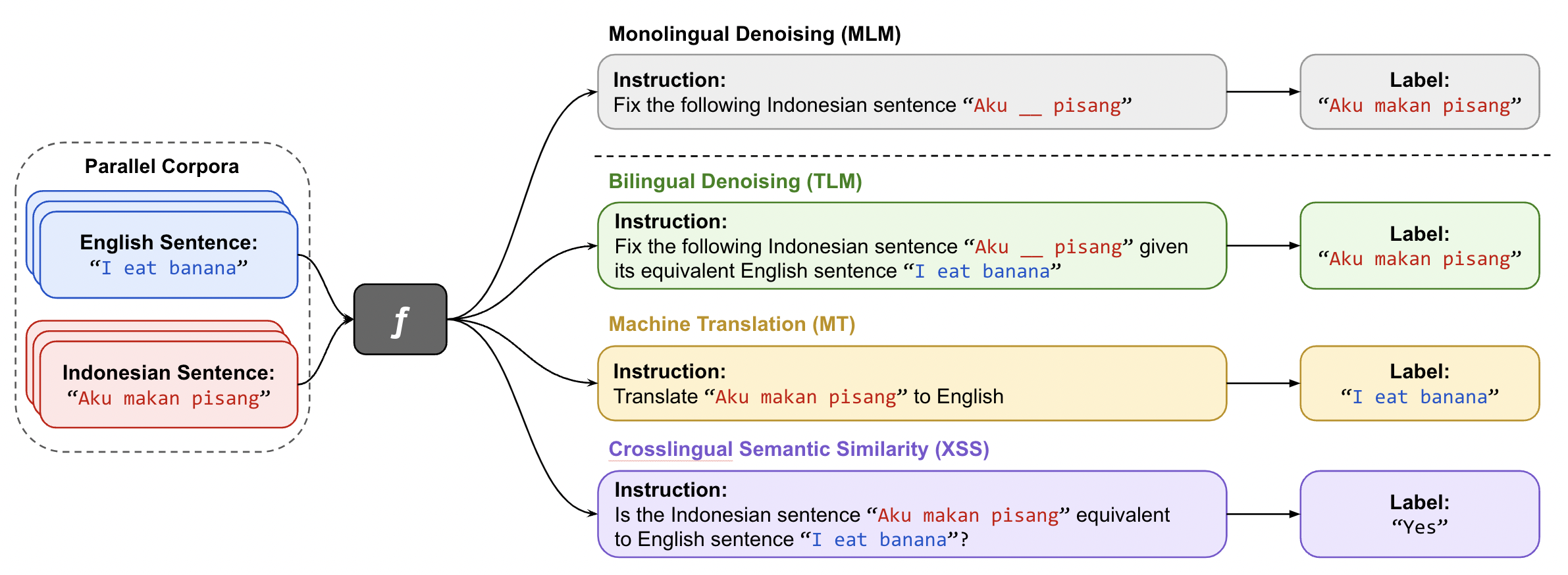}
    }
    \caption{Example of the alignment-based crosslingual instruction prompts, i.e., bilingual denoising (TLM), machine translation (MT), and crosslingual semantic similarity (XSS) in comparison to the monolingual denoising (MLM).} 
    \label{fig:prompt-example}
\end{figure*}


\footnotetext{We gather the language and language family information from URIEL~\cite{littell2017uriel,malaviya17emnlp}.}

In summary, our work presents the following major contributions: 

\begin{itemize}
    \item We propose InstructAlign, a crosslingual continual instruction tuning method that allows instruction-tuned LLMs to understand \textcolor{teal}{L2} with minimal degradation on \textcolor{red}{L1} while retaining their zero-shot prompting capability.
    \item We propose alignment-based crosslingual instruction tuning, which enables LLMs to align \textcolor{teal}{L2} to \textcolor{red}{L1} allowing better \textcolor{teal}{L2} acquisition with only a limited amount of parallel data.
    \item We evaluate the effectiveness of InstructAlign on Indonesian local languages datasets, and demonstrate that InstructAlign can significantly improve the performance on \textcolor{teal}{L2} by 5-10\% F1 while maintaining the original performance on \textcolor{red}{L1} and its multitask capability.
    \item We analyze the correlation between the performance of \textcolor{teal}{L2} and other unseen languages (\textcolor{violet}{L3}), suggesting the zero-shot generalization of InstructAlign to \textcolor{violet}{L3} particularly when the languages are related.~\footnote{We use the terms \textcolor{red}{L1}, \textcolor{teal}{L2}, and \textcolor{violet}{L3} to denote the first, second, and third language acquisition~\cite{hammarberg2001lang-acq,hammarberg2014lang-acq}. In our context, \textcolor{red}{L1} denotes the pre-trained languages in LLMs, \textcolor{teal}{L2} denotes the newly adapted languages, and \textcolor{violet}{L3}denotes other languages that have not been seen after tuning with InstructAlign, which are only used in the evaluation.}
\end{itemize}

\section{Related Work}

\subsection{Instruction Tuning in LLMs}

Early works~\cite{wei2022flan,chung2022flan-t5,sanh2022t0,ouyang2022instructgpt} have shown the effectiveness of instruction-tuned LLMs, which significantly improves the zero-shot generalization capability over the corresponding non-instruction-tuned LLMs by a huge margin. Since then, various instruction-tuned LLMs have been released, including T0~\cite{sanh2022t0}, InstructGPT~\cite{ouyang2022instructgpt}, FLAN-GPT~\cite{wei2022flan}, FLAN-T5~\cite{chung2022flan-t5}, FLAN-PaLM~\cite{chung2022flan-t5}, mT0~\cite{muennighoff2022crosslingual}, BLOOMZ~\cite{muennighoff2022crosslingual}, Alpaca~\cite{taori2023alpaca}, etc. However, most of these models are only pre-trained on a single or few languages, with the exception of mT0 and BLOOMZ which are adapted from models pre-trained on 101 languages, i.e., mT5~\cite{xue2021mt5}, and pre-trained on 46 languages, i.e., BLOOM~\cite{scao2022bloom}, respectively. In this work, we utilize BLOOMZ~\cite{muennighoff2022crosslingual} as the backbone in of InstructAlign.

\subsection{Crosslingual Alignment}

Crosslingual alignment is a widely explored concept that allows language models (LMs) to align, commonly at a word/sentence level, across different languages. Crosslingual alignment allows the models to perform crosslingual inference without requiring any tuning on the target task. \citet{fung1997nonparallel,fung1998nonparallel} a bilingual lexicon extraction method through word-to-word alignment from word relation matrix. 
\citet{fung2004nonparallel} introduces a bilingual lexicon and parallel sentence extraction method from aligning sentences from non-parallel data via Bootstrapping and EM.
\citet{lample2018word,cao2020multilingual} introduces align bilingual lexicon method that requires no parallel data by performing embedding alignment across different languages. This is then utilized to deal with unsupervised machine translation \citet{lample2018unsupervised}. 
A crosslingual pre-training objective for building LMs, namely translation language modeling (TLM)~\cite{conneau2019xlm}, has also been explored which enforces token-level alignment between languages allowing the model to learn aligned representation across multiple languages. In this work, we perform crosslingual alignment through instruction by introducing bilingual denoising instruction which is equivalent to token-level alignment in TLM, and translation instruction which serves as sentence-level alignment across different languages.

\subsection{Continual Learning for Language Models}

Continual learning is a paradigm to learn various tasks gradually allowing the model to acquire new knowledge over time\cite{delnage2021continual}. Using a naive fine-tuning approach for continual learning causes the model to suffer from catastrophic forgetting (CF)~\cite{french1999cf}. Therefore, various methods have been introduced to prevent CF. Regularization-based methods~\cite{kirkpatrick2017ewc,liu2018rewc,aljundi2018mas} add a regularization in the loss function to prevent the model to be updated into a direction that causes CF. Replay-based methods~\cite{david2019experiencereplay,david2017gem,chaudhry2019agem} add samples from previous tasks to be incorporated during learning the new task, which helps regularize the model to avoid CF. Parameter isolation methods~\cite{aljundi2017expertgate,serra2018hat,mallya2018packnet} prevent the model from CF by learning new tasks using a new set of parameters while keeping the other parameters frozen during fine-tuning. In this work, we apply experience replay~\cite{david2019experiencereplay}, which is a simple replay-based method by adding tasks from previously learned languages when training new languages without any loss modification. 

\section{Methodology}

\begin{table*}[!t]
    \centering
    \resizebox{0.73\linewidth}{!}{
    \begin{tabular}{llccccc}
    \toprule
    \multicolumn{1}{c}{\textbf{Dataset}} & \multicolumn{1}{c}{\textbf{Task}} & \textbf{\#Lang.} & \textcolor{red}{\textbf{\#L1}}  & \textcolor{teal}{\textbf{\#L2}} & \textcolor{violet}{\textbf{\#L3}} & \textbf{\#Test} \\
    \midrule
    NusaX           &Sentiment Analysis   &12 &2 &7 &3 &4400 \\
    NusaTranslation &Sentiment Analysis   &11+1 &1 &3 &8 &10400 \\
    NusaParagraph   &Emotion Recognition  &10 &0 &4 &6 &5700 \\
    NusaParagraph   &Topic Classification &10 &0 &4 &6 &6250 \\
    \bottomrule
    \end{tabular}
    }
    \caption{Statistics of all datasets used in the experiments. \textbf{\#Lang.} denotes the \#languages in each dataset.}
    \label{tab:datasets}
\end{table*}

InstructAlign is a continual crosslingual instruction tuning framework that allows the model to align high-to-low resource languages through instruction tuning. InstructAlign introduces two components, i.e., 1) crosslingual alignment through instruction tuning, which allows the model to align pre-trained languages with the new languages through crosslingual alignment, and 2) continual instruction tuning, which applies continual learning into instruction tuning to avoid catastrophic forgetting.

\subsection{Crosslingual Alignment through Instruction}

Given a parallel text pair $(x, y)$ from two languages, the goal of crosslingual alignment is to learn a mapping function $f(.)$ parameterized by $\theta$ such that $f(x, \theta) = f(y, \theta)$. The $(x,y)$ text pair commonly comes in the form of a word pair or a phrase pair~\cite{lample2018word,lample2018unsupervised}, but in theory, it should be able to generalize to a sentence pair or even a paragraph. With the goal of aligning two parallel texts from two different languages, InstructAlign defines a set of alignment-based crosslingual instructions by exploiting multiple alignment objectives that can be achieved through a parallel sentence. Specifically, we explore three different objectives, i.e., bilingual denoising / translation language modeling (\textbf{TLM}), machine translation (\textbf{MT}) and crosslingual semantic similarity (\textbf{XSS}). 

We first define a parallel sentence pair $(X=\{x_1,x_2,\dots,x_m\}, Y=\{y_1,y_2,\dots,y_n\})$, where $x_i$ and $y_i$ denote the $i$-th token of the sentence $X$ and $Y$, respectively. For bilingual denoising (\textbf{TLM}), we model the problem as a conditional denoising task. InstructAlign first applies a perturbation function $g^{tlm}(.)$ to the target sentence $Y$ that masks out part of the tokens in order to get $\tilde{Y} = g^{tlm}(Y)$. The pair $(X, \tilde{Y})$ is then used to generate a prompt using $h(X,\tilde{Y},T^{tlm})$, resulting in an input-output data pair for prompting $(h^{tlm}(X,\tilde{Y},T^{tlm}), Y)$, where $h^{tlm}(.)$ denotes a bilingual denoising prompt generator and $T^{tlm}$ the prompt template. 

For the machine translation (\textbf{MT}) objective, we define the input-output data pair as $(h^{mt}(X,T^{mt}), Y)$, where $h^{mt}(.)$ denotes a machine translation prompt generator and $T^{mt}$ denotes a machine translation prompt template. As for the crosslingual semantic similarity (\textbf{XSS}) objective, we models the problem as an inference task to predict whether two parallel sentences X and Y are semantically similar. Specifically, we define the input-output data pair as $(h^{xss}(X,Y,T^{xss}), l)$ where $h^{xss}(.)$ is a semantic similarity prompt generator, $T^{xss}$ denotes a semantic similarity prompt template and $l$ the binary label regarding whether the sentences are semantically related or not. The examples of the crosslingual alignment objectives are shown in Figure \ref{fig:prompt-example}.

\begin{table*}[!t]
    \centering
    \resizebox{0.95\linewidth}{!}{
    \begin{tabular}{l|c|c|c|c|c|c|c|c|c}
    \toprule
        \multicolumn{1}{c|}{\multirow{2}{*}{\textbf{Method}}} & \multicolumn{5}{c|}{\textbf{\textcolor{teal}{L2} Weighted F1 (\%)}} & \multicolumn{4}{c}{\textbf{\textcolor{red}{L1} Weighted F1 (\%)}} \\
        \cmidrule{2-10}
        & \textbf{NT-S} & \textbf{NX-S} & \textbf{NP-E} & \textbf{NP-T} & \textbf{Avg.} & \textbf{NX-S En} & \textbf{NX-S Id} & \textbf{NT-S Id} & \textbf{Avg.} \\

    \midrule
       \multicolumn{10}{c}{\textbf{BLOOM \& BLOOMZ Baseline}} \\
    \midrule
        \textbf{BLOOM-560M} & 57.62 & 21.80 & \underline{2.80} & 5.34 & 21.89 & 29.26 & 21.13 & 61.47 & 37.29 \\
        \textbf{BLOOM-1.1B} & 59.18 & 22.02 & \underline{2.80} & 5.35 & 22.34 & 22.02 & 22.54 & 58.81 & 34.46 \\
        \textbf{BLOOM-3B} & 44.98 & 21.21 & \underline{2.80} & 5.35 & 18.59 & 24.03 & 21.17 & 58.30 & 34.50 \\
    \midrule
        \textbf{BLOOMZ-560M} & 46.83 & 33.73 & \underline{2.80} & 5.35 & 22.18 & 58.24 & 55.59 & 69.81 & 61.21 \\
        \textbf{BLOOMZ-1.1B} & 64.01 & 41.50 & \underline{2.80} & 5.35 & 28.42 & 57.41 & 58.58 & 80.40 & 65.46 \\
        \textbf{BLOOMZ-3B} & \underline{69.41} & \underline{45.82} & \underline{2.80} & \underline{5.73} & \underline{30.94} & \underline{62.65} & \textbf{\underline{63.21}} & \textbf{\underline{81.38}} & \textbf{\underline{69.08}} \\
    \midrule
       \multicolumn{10}{c}{\textbf{InstructAlign-Tuned BLOOMZ-560M}} \\
    \midrule
        \textbf{MLM r=100k} & 66.51 & 42.51 & 2.80 & 5.52 & 29.34 & 60.97 & \underline{60.01} & 70.93 & 63.97 \\
        \textbf{MT Obj. r=100k} & 66.42 & 41.20 & 2.82 & 5.40 & 28.96 & 60.96 & 58.09 & 64.18 & 61.08 \\
        \textbf{TLM r=100k} & \underline{69.24} & 42.91 & \underline{2.87} & 5.43 & 30.11 & 61.65 & 58.52 & \underline{72.40} & \underline{64.19} \\
        \textbf{XSS r=100k} & 68.10  & \underline{45.83} & 2.84 & \underline{5.53} & \underline{30.58} & \underline{61.89} & 58.22 & 71.27 & 63.79 \\
    \midrule
       \multicolumn{10}{c}{\textbf{InstructAlign-Tuned BLOOMZ-1.1B}} \\
    \midrule  
        \textbf{MLM r=100k} & 71.46 & 45.73 & 2.84 & 5.49 & 31.38 & 61.30 & \underline{60.83} & 73.25 & 65.13 \\
        \textbf{MT Obj. r=100k} & 66.15 & 44.93 & 2.84 & 5.40 & 29.91 & 61.68 & 59.18 & 65.28 & 62.05\\
        \textbf{TLM r=100k} & 70.29 & \textbf{\underline{49.25}} & \textbf{\underline{3.17}} & \textbf{\underline{6.34}} & 32.26 & 63.26 & 60.54 & 74.66 & 66.15\\
        \textbf{XSS r=100k} & \textbf{\underline{71.89}} & 49.23 & 3.08 & 5.81 & \textbf{\underline{32.50}} & \textbf{\underline{63.78}} & 59.34 & \underline{75.74} & \underline{66.29} \\
    \bottomrule
    \end{tabular}
    }
    \caption{Evaluation results of InstructAlign with BLOOMZ-560M and BLOOMZ-1.1B backbones. Compared to BLOOM and BLOOMZ baselines, All InstructAlign-tuned models improve the zero-shot crosslingual performance in \textcolor{teal}{L2} while also retaining the performance in \textcolor{red}{L1}. 
    }
    \label{tab:evaluation-result}
\end{table*}

\subsection{Continual Instruction Tuning through Experience Replay}

Within the continual instruction tuning phase of InstructAlign, experience replay~\cite{david2019experiencereplay} is employed to minimize the catastrophic forgetting problem. mhamdi-etal-2023-cross Experience replay works by storing some of the past training data and using them during the optimization step of the new data. These past data serve as a regularization term that prevents the models to forget past knowledge when learning from the new data. The past data is collected from the instruction tuning data used when developing the corresponding instruction-tuned model, which are all supervised. 

During the continual instruction tuning, InstructAlign takes only $r$ randomly sampled data from the past instruction tuning data. The sampled past data is used during continual-instruction tuning with a balanced sampling between the past data and new data. More formally, we define a past dataset $\mathcal{D}^{old}$ and a newly generated crosslingual instruction dataset $\mathcal{D}^{cli}$. On each optimization step, InstructAlign samples data in an interleaving manner resulting in a batch data $\mathcal{B}=\{s^{\mathcal{D}^{old}}_1, s^{\mathcal{D}^{cli}}_1, s^{\mathcal{D}^{old}}_2, s^{\mathcal{D}^{cli}}_2, \dots, s^{\mathcal{D}^{old}}_n, s^{\mathcal{D}^{cli}}_n\}$ with $2n$ samples, where $s^{\mathcal{D}^{old}}_i$ and $s^{\mathcal{D}^{cli}}_i$ denote a sample that is taken randomly from $\mathcal{D}^{old}$ and $\mathcal{D}^{cli}$, respectively. Since the samples are all supervised, the optimization can be done by optimizing the cross-entropy loss~\cite{good1952ce} from all the samples in the batch.

\section{Experiment Setting}

\subsection{Continual-Instruction Tuning Dataset} 

During the InstructAlign tuning, we train the model on 7 \textcolor{teal}{L2} languages from Malayo-Polynesian language family group, i.e., Sundanese (sun), Javanese (jav), Balinese (ban), Minangkabau (min), Buginese (bug), Acehnese (ace), and Banjarese (bjn). For the \textcolor{red}{L1} languages, we utilize English (eng), as English covers the majority of the pre-training data in most LLMs, and Indonesian (ind), as the language is closely related to the target \textcolor{teal}{L2} languages. For the dataset, we utilize FLORES-200 dataset~\cite{goyal2021flores101,nllb2022nllb} as the source of the parallel data where we combine the validation and the test set producing a total of $\sim$2000 parallel sentences for each language pair which is orders of magnitude smaller compared the data size used for language adaptation~ used in prior works\cite{pfeiffer2020mad,cahyawijaya2021indonlg,alabi2022adapting,yong2022bloom1}.

\subsection{Models \& Hyperparameters} 

We utilize BLOOMZ~\cite{muennighoff2022crosslingual} as the backbone model. Specifically, we explore InstructAlign on two model size, i.e., BLOOMZ-560M and BLOOMZ-1.1B. For InstructAlign, we evaluate three crosslingual alignment objectives, i.e., \textbf{TLM}, \textbf{XSS}, and \textbf{MT}. The list of prompts used for instruction tuning is described in Appendix~\ref{app:prompts}. We use English prompts in all experiments. We run all experiments with an initial learning rate of 1e-5 with a linear learning rate decay and a batch size of 32 for a fixed optimization step of 50,000. We run the InstructAlign on a single RTX3090 GPU (24GB) using the AdamW optimizer~\cite{loshchilov2018decoupled} and mixed-precision training~\cite{micikevicius2018mixed}. We use a fixed number of replay samples $r$ = 100000.

\subsection{Evaluation Setting}

\label{sec:evaluation-setting}

After tuning with InstructAlign, the model is then evaluated in a zero-shot crosslingual inference setting, in which the model has never seen the task on the target languages, but might have seen the task on other seen languages. To retrieve the classification label, we compute the joint probability of the prompt with each label in the dataset and pick the label which prompt the highest joint probability. We consider 3 different prompts in English for the zero-shot inference and take the average accuracy and weighted F1 scores as the evaluation metrics.  The list of the prompts used in our evaluation is shown in Appendix~\ref{app:prompts}. We use a single RTX1080Ti GPU (11GB) to run the evaluation for all models. To reduce the memory bottleneck during inference, we run the evaluations using 8-bit inference via LLM.int8()~\cite{dettmers2022llmint8}. We provide the performance comparison between 8-bit and 32-bit evaluation in Appendix~\ref{app:8bit-vs-32bit}.

\begin{figure}
    \centering
    \resizebox{\linewidth}{!}{
    \includegraphics{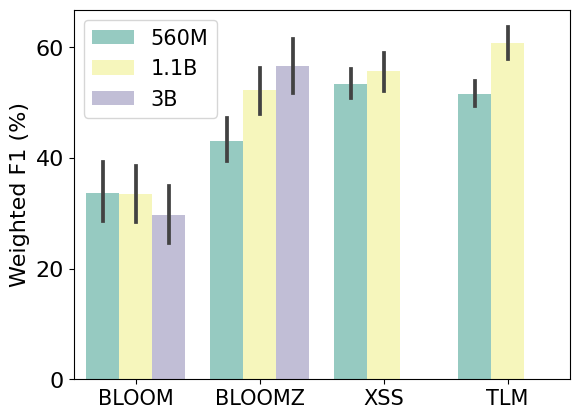}
    }
    \caption{Average performance of various models across different model scales on the \textcolor{red}{L1} and \textcolor{teal}{L2} languages subsets of the \textbf{NT-S} and \textbf{NX-S} datasets.}
    \label{fig:perf-scaling}
    \vspace{-5pt}
\end{figure}

\paragraph{Zero-Shot Evaluation Datasets} For evaluating the effectiveness of InstructAlign, we utilize four multilingual Indonesian local languages datasets, i.e., the sentiment analysis task from NusaX (\textbf{NX-S})~\cite{winata2022nusax}, the sentiment analysis task from NusaTranslation (\textbf{NT-S})~\cite{cahyawijaya2023nusawrites}, the topic classification task from NusaParagraph (\textbf{NP-T})~\cite{cahyawijaya2023nusawrites}, and the paragraph-level emotion recognition task from NusaParagraph (\textbf{NP-E})~\cite{cahyawijaya2023nusawrites}. The detailed per-dataset statistics are shown in Table~\ref{tab:datasets}.
NusaX covers 12 languages including 2 \textcolor{red}{L1} languages: English (eng) and Indonesian (ind), 7 \textcolor{teal}{L2} languages: Acehnese (ace), Balinese (ban), Buginese (bug), Banjarese (ban), Javanese (jav), Minangkabau (min), and Sundanese (sun), and 3 \textcolor{violet}{L3} languages: Toba Batak (bbc), Madurese (mad), and Ngaju (nij). While NusaTranslation covers 11 languages, which includes 3 \textcolor{teal}{L2} languages: Javanese (jav), Sundanese (sun), and Minangkabau (min), and 8 \textcolor{violet}{L3} languages: Ambon (abs), Batak (btk), Betawi (bew), Bima (bhp), Madurese (mad), Makassarese (mak), Musi (mui), and Rejang (rej). NusaParagraph covers 10 languages, which includes 4 \textcolor{teal}{L2} languages: Sundanese, Javanese (jav), Minangkabau (min), and, and 6 \textcolor{violet}{L3} languages: Batak (btk), Betawi (bew), Madurese (mad), Makassarese (mak), Musi (mui), Rejang (rej). To expand the evaluation dataset for \textcolor{red}{L1}, we add the Indonesian sentiment analaysis data from IndoLEM~\cite{koto2020indolem}~\footnote{The source translation data of NusaTranslation} as the Indonesian (ind) subset of \textbf{NT-S}. More details about each dataset can be found in Appendix~\ref{app:datasets}.

\subsection{Baselines}

For our baselines, we conduct zero-shot prompting using four different sizes of BLOOMZ, i.e., BLOOMZ-560M, BLOOMZ-1.1B, BLOOMZ-1.7B, and BLOOMZ-3B, without any additional language adaptation phase. In addition, to compare the effectiveness of the crosslingual alignment, we add continual instruction-tuned baselines that incorporate only monolingual denoising instructions, which is equivalent to performing language adaptation using MLM~\cite{devlin2019bert}. 

\section{Experiment Result}
\label{sec:result}

\begin{table}[!t]
    \centering
    \resizebox{\linewidth}{!}{
    \begin{tabular}{l|c|c}
    \toprule
    \textbf{Method} &\textbf{\textcolor{teal}{L2}} & \textbf{\textcolor{red}{L1}} \\
    \midrule
    \multicolumn{3}{l}{\qquad \textit{Baselines}} \\
    \midrule    
    Random                                       & \underline{40.28} & 30.88 \\
    Majority                                     & 32.34 & 21.17 \\
    BLOOMZ-560M                                  & 37.66 & \textbf{61.21} \\
    \midrule
    \multicolumn{3}{l}{\qquad \textit{Single Objective}} \\
    \midrule    
    Monolingual Denoising (MLM)                  & 36.71 & 53.14 \\
    Machine Translation (MT)	                 & 35.43 & 47.95 \\
    Bilingual Denoising (TLM)	                 & 45.48 & 53.28 \\
    Crosslingual Semantic Similarity (XSS)	     & 44.55 & 54.05 \\
    \midrule
    \multicolumn{3}{l}{\qquad \textit{Multi Objectives}} \\
    \midrule
    MLM + MT & 40.09 & 47.67 \\
    TLM + MT & 42.93 & 48.75 \\
    XSS + MT & 43.32 & 50.66 \\
    MLM + TLM & 43.46 & 53.16 \\
    MLM + XSS & 42.82 & 53.90 \\
    TLM + XSS & \textbf{\underline{45.83}} & \underline{54.01} \\
    \bottomrule
    \end{tabular}
    }
    \caption{Averaged Weighted F1 scores from various InstructAlign objectives in the \textbf{NT-S} and \textbf{NX-S} datasets. We use BLOOMZ-560M as the backbone.}
    \label{tab:alignment-objectives}
\end{table}

\begin{figure*}[!t]
\centering
\begin{subfigure}{0.45\linewidth}
\centering
  \includegraphics[width=\linewidth]{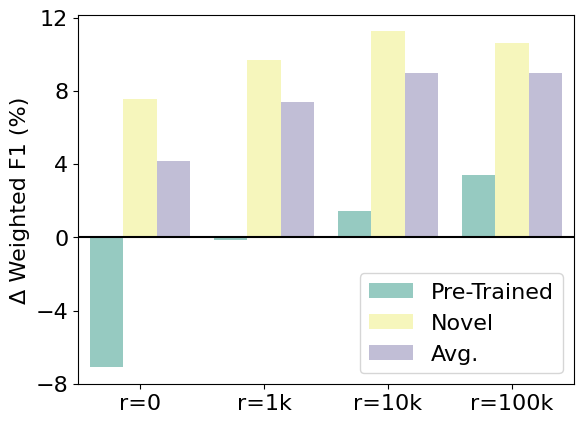}
\end{subfigure}
\hspace{1em}
\begin{subfigure}{0.45\linewidth}
\centering
  \includegraphics[width=\linewidth]{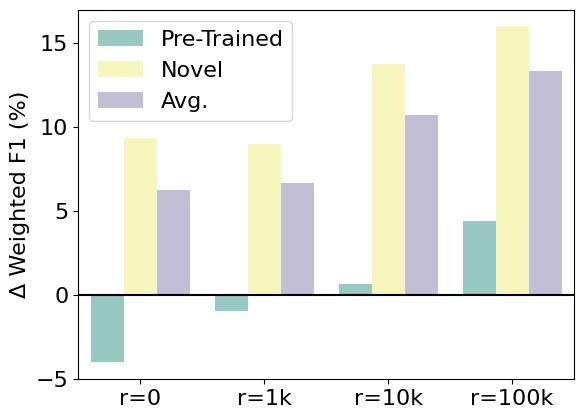}
\end{subfigure}
\vspace{-4pt}
\caption{$\Delta$ weighted F1 of InstructAlign tuned BLOOMZ-560M with \textbf{(left)} TLM  and \textbf{(right)} XSS objectives various continual instruction-tuned approaches compared to the original BLOOMZ-560M baseline. Negative scores indicate that the model performs worse compared to the baseline.}
\label{fig:delta-per-lang}
\end{figure*}


\paragraph{Effectiveness of InstructAlign} Table~\ref{tab:evaluation-result} shows the result of InstructAlign on both \textcolor{red}{L1} and \textcolor{teal}{L2} languages. InstructAlign-tuned models with MT, TLM, and XSS objectives significantly outperform the comparable-sized BLOOM and BLOOMZ baselines on \textcolor{teal}{L2} languages while retaining a similar performance level as the original BLOOMZ models on \textcolor{red}{L1} languages. Surprisingly, InstructAlign with MLM objectives is also effective, yielding a similar performance on \textcolor{teal}{L2} languages compared to the crosslingual objectives. In $\S$~\ref{sec:alignment-obj}, we show that this improvement only occurs after combining the MLM objective with the experience replay, demonstrating the importance of continual instruction tuning during language adaption.
While in the NusaParagraph emotion recognition (\textbf{NP-E}) and topic classification (\textbf{NP-T}) tasks, all baselines yield a very low score, suggesting that the ability to solve long text classification tasks do not emerge on that scale~\cite{wei2022emergent}. Interestingly, InstructAlign tuned models indicate consistent improvement, although marginal, on these tasks, demonstrating that an early emergence in \textcolor{teal}{L2} languages is possible through InstructAlign.

\paragraph{Effect of Model Scaling} As shown in Figure~\ref{fig:perf-scaling}, we observe that scaling increases the zero-shot performance of BLOOMZ on both \textcolor{red}{L1} and \textcolor{teal}{L2}, but the same does not apply to BLOOM, suggesting the benefit of instruction tuning for better generalization to unseen tasks and languages. Moreover, applying InstructAlign on larger BLOOMZ results in higher overall zero-shot performance on both \textcolor{red}{L1} and \textcolor{teal}{L2}. Specifically, InstructAlign-tuned models with 1.1 billion parameters yield $\sim$2\% higher performance compared to the 560 million parameters InstructAlign-tuned models and even perform competitively with the original 3 billion parameters BLOOMZ model. This suggests that the scaling law of language models~\cite{kaplan2020scaling,hoffmann2022chinchila} also apply after InstructAlign where larger-sized models tend to perform better compared to their smaller counterpart. Detailed experiment results are described in Appendix~\ref{app:results}.

\section{Analysis and Discussion}

\subsection{Alignment Objectives}
\label{sec:alignment-obj}

To better understand the effectiveness of each alignment objective, we conduct experiments by using a single objective, i.e., monolingual denoising (MLM), machine translation (MT), bilingual denoising (TLM) and crosslingual semantic similarity (XSS), as well as multi objectives on various combinations. We also test zero-shot prompting without any additional language adaption phase as a baseline for comparison. Note that continual instruction tuning through experience replay is not applied ($r$=0) in these experiments since we focused on the effect of alignment objectives.

As shown in Table~\ref{tab:alignment-objectives}, BLOOMZ 560M zero-shot performs better than the random baseline on \textcolor{red}{L1} while achieving a lower score on \textcolor{teal}{L2}, showing that BLOOMZ 560M is unable to be directly applied to these \textcolor{teal}{L2} languages. For InstructAlign with a single objective, similar to the result from prior work~\cite{yong2022bloom1}, applying the MLM objective decays the performance of the model. Similarly, using MT objective also decreases the performance of both \textcolor{red}{L1} and \textcolor{teal}{L2}. Nevertheless, as shown in Table~\ref{tab:evaluation-result}, this problem can be mitigated by applying continual learning. On the other hand, both TLM and XSS help improve the model on \textcolor{teal}{L2}, indicating that these objectives are effective for aligning \textcolor{red}{L1} and \textcolor{teal}{L2} languages. Additionally, the performance in \textcolor{red}{L1} languages is also retained the most when using the TLM and XSS objectives.

When combining multiple objectives during InstructAlign, we observe the highest score when combining TLM and XSS. Interestingly, adding the MLM and MT objectives during InstructAlign consistently yields a lower score compared to the single TLM and XSS objectives for both \textcolor{teal}{L2} and \textcolor{red}{L1} languages. These facts suggest that cross-lingual objectives such as XSS and TLM, are effective for learning new languages through cross-lingual instruction-tuning with limited data.



\subsection{Continual Instruction Tuning}

In order to assess the effectiveness of continual instruction tuning through experience replay, we conduct an experiment exploring the effect of different numbers of replay samples $r$ used in continual instruction tuning. Specifically, we explore 4 settings of $r$, i.e., $r=[0, 1000, 10000, 100000]$. Figure~\ref{fig:delta-per-lang} shows the performance of the InstructAlign tuned models across different ranges of replay examples $r$. When using no experience replay ($r$=0), the performance of the \textbf{pre-trained} languages drops significantly, and even further, the performances on the \textbf{novel} languages also drop which suggests that the multitask prompting capability for both of these methods are degraded~\cite{yong2022bloom1}. When $r$ increases, a much smaller performance degradation is observed on the \textcolor{red}{L1} languages. Interestingly, the performance on \textbf{novel} languages also improved when $r$ increases which in the end, increases the performance of the model across all languages. These facts demonstrate the importance of the experience replays for avoiding catastrophic forgetting in continual instruction tuning.




\subsection{Impact of InstructAlign on \textcolor{violet}{L3} Languages}

\begin{figure}[!t]
    \centering
    \includegraphics[width=\linewidth]{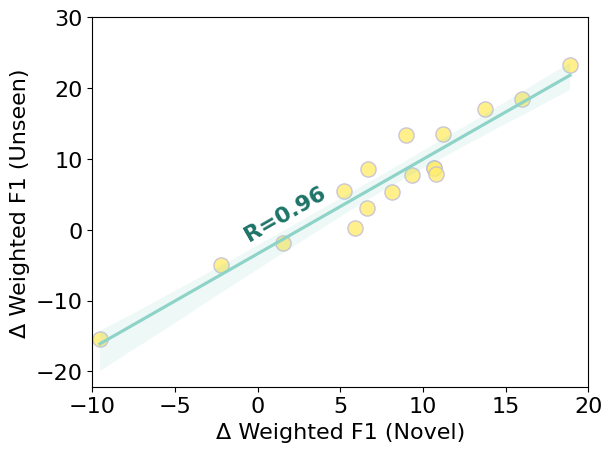}
    \vspace{-4pt}
\caption{Correlation of $\Delta$ weighted F1 from the InstructAlign tuned models to the corresponding BLOOMZ backbone models on novel and unseen languages. $R$ denotes the Pearson correlation coefficient.}
\label{fig:corr-unseen-lang}
\end{figure}

We further assess the impact of aligning \textcolor{teal}{L2} languages through InstructAlign to other unseen Indonesian languages which are within the same language family group (L3). To assess the effectiveness of transferability from the \textcolor{teal}{L2} languages to \textcolor{violet}{L3} languages, we compute the correlation coefficient between $\Delta$ weighted F1 score on the \textcolor{teal}{L2} and \textcolor{violet}{L3} languages for each model compared to the corresponding baseline, and measure the Pearson's correlation coefficient~\cite{rodgers1988correlation,freedman2007statistics}.

\begin{figure}[!t]
    \centering
    \includegraphics[width=\linewidth]{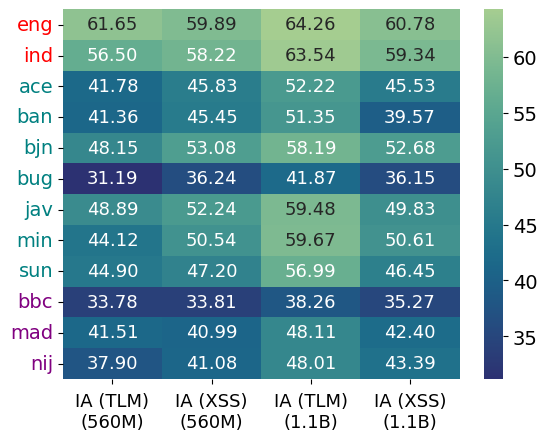}
    \vspace{-5pt}
\caption{Per language results of InstructAlign tuned models in NusaX. \textcolor{red}{red} denotes \textcolor{red}{L1} languages. \textcolor{teal}{teal} denotes \textcolor{teal}{L2} languages. \textcolor{violet}{purple} denotes \textcolor{violet}{L3} languages}
\label{fig:unseen-per-lang}
\end{figure}

As shown in Figure~\ref{fig:corr-unseen-lang}, the correlation coefficient between the performance improvement of \textcolor{teal}{L2} and \textcolor{violet}{L3} languages is high with a Pearson's correlation coefficient of 0.96. This indicates the effectiveness of the InstructAlign approach for not only adapting to \textcolor{teal}{L2} languages but also to related \textcolor{violet}{L3} languages. Nevertheless, the improvement for unseen language still depends on the language distances as shown in Figure~\ref{fig:unseen-per-lang}, where performance on Toba Batak (bbc) and Buginese (bug) yield much lower scores compares the other languages. This result aligns with the analysis from NusaX~\cite{winata2022nusax} which shows that the performances of Buginese (bug) and Toba Batak (bbc) are the lowest for both the multitask and zero-shot crosslingual settings due to the relatively low vocabulary overlapping compared to other languages in NusaX. This suggests that by performing , the model can also understand unseen languages that are related to the novel-adapted language, indicating the generalization of the crosslingual transfer from pre-trained languages to novel and unseen languages


\subsection{Conclusion}

In this work, we address the challenge of increasing the language coverage of instruction-tuned LLMs by introducing a crosslingual continual instruction tuning method, InstructAlign. We demonstrate that InstructAlign allows an instruction-tuned LLM to effectively learn novel languages through alignment-based crosslingual instruction tuning objectives while retaining the existing multitask and multilingual abilities. Based on our experiment results on four Indonesian local languages datasets, InstructAlign effectively improves the understanding of novel Indonesian local languages, improving the language understanding performance on novel languages by $\sim$5-10\% weighted F1 score and also demonstrates a better forward transfer performance to other unseen Indonesian local languages by a significant margin. In addition, we analyze various objectives of InstructAlign and demonstrate the effectiveness of alignment-based crosslingual instruction tuning objectives compared to the traditional masked language modeling (MLM) for learning novel languages with a limited amount of data. Our work contributes to the advancement of language adaptation methods for instruction-tuned LLMs, especially for underrepresented languages.

\section{Limitation and Future Works}

\subsection{Other Model Architectures}

Despite the effectiveness of InstructAlign on BLOOMZ, its effectiveness has not been explored for different model architectures such as encoder-decoder or other model architectures. Due to the limited computing budget, we can only run the InstructAlign experiment on a decoder-only model, i.e., BLOOMZ, We encourage future works to explore the experiment in other model architectures. 

\subsection{Scaling to Larger LLMs}

As described in $\S$\ref{sec:result}, we hypothesize that InstructAlign-tuned models follow the scaling laws of language models~\cite{kaplan2020scaling,hoffmann2022chinchila}. Nevertheless, we can only empirically show this scaling effect of InstructAlign in BLOOMZ-560M and BLOOMZ-1.1B due to the limited computing budget. We expect future works to expand the exploration to larger-scale models.

\subsection{Other Continual Learning Methods}

In terms of continual learning methods, we only explore a single approach, i.e., experience replay~\cite{david2019experiencereplay}, due to the efficient memory requirement of this method. Further analysis and examination of other potential continual learning approaches, such as A-GEM~\cite{chaudhry2019agem} and EWC~\cite{liu2018rewc}, is another potential research direction to be explored in future works.

\subsection{Underrepresented Languages from Other Language Family}

There are many other underrepresented languages such as indigenous languages of the Americas~\cite{ebrahimi2022americasnli}, African~\cite{adelani2021masakhaner,adelani2022masakhaner2}, Indic~\cite{kakwani2020indicnlpsuite,kumar2022indicnlg}, Austronesian~\cite{winata2022nusax,cahyawijaya2023nusacrowd}, and many others all around the world. In this work, we only explore InstructAlign for Malayo-Polynesian language family group under the Austronesian language family, specifically for Indonesian local languages. For future work, we are eager to explore the generalization of InstructAlign and other language adaptation methods on other underrepresented and low-resource languages.

\section*{Ethical Consideration}

Our work highlights the importance of inclusivity in LLM technology for underrepresented and extremely low-resource languages. During our study, we are well aware of the ethical responsibility associated with language research and the potential impact it can have on communities. Our ultimate goal is to promote linguistic diversity and contribute to a more inclusive NLP landscape. We encourage further collaboration and engagement with underrepresented and low-resource language communities to ensure that their voices are heard and their needs are addressed in future language technology development. We remain committed to the principles of ethical research, diversity, inclusivity, and fairness, striving to mitigate biases and promote social good through our work in the field of NLP.

\section{Acknowledgements}

We thank Bryan Wilie and Yong Zheng Xin for the fruitful discussions and suggestions.  This work has been partially funded
by PhD Fellowship Award, the Hong Kong University of Science and Technology; and PF20-43679 Hong Kong PhD Fellowship Scheme, Research Grant Council, Hong Kong.

\clearpage

\bibliography{anthology,acl_latex}
\bibliographystyle{acl_natbib}

\clearpage

\appendix

\section{Prompt List}
\label{app:prompts}

In this section, we provide the list of the prompt used in our experiment. For InstructAlign, we use 6 prompts for each objective. The prompt list for bilingual denoising (\textbf{TLM}), machine translation (\textbf{MT}), crosslingual semantic similarity (\textbf{XSS}), and monolingual denoising (\textbf{MLM}) are shown in Table~\ref{tab:tlm-prompt}, Table~\ref{tab:mt-prompt}, Table~\ref{tab:xss-prompt}, and Table~\ref{tab:mlm-prompt}, respectively. For the evaluation, we employ 3 English prompts for each task. The prompt list for sentiment analysis, emotion recognition, and topic classification tasks are described in Table~\ref{tab:senti-prompt}, Table~\ref{tab:emot-prompt}, and Table~\ref{tab:topic-prompt}, respectively.

\begin{table*}[!t]
    \centering
    \resizebox{0.9\linewidth}{!}{
        \begin{tabular}{p{\linewidth}}
            \toprule
            \textbf{Prompt in \colorbox{yellow!25}{Bilingual Denoising (TLM)} Task}  \\
            \midrule
            \texttt{[INPUT\_TEXT]. Denoise the previous [INPUT\_LANG] text to its equivalent sentence in [CONTEXT\_LANG]: [CONTEXT]\textbackslash n[LABEL\_TEXT]} \\
            \midrule
            \texttt{Context in [CONTEXT\_LANG]: [CONTEXT]\textbackslash nFix the following [INPUT\_LANG] text "[INPUT\_TEXT]" ensuring the meaning is equivalent with the context. [LABEL\_TEXT]} \\
            \midrule
            \texttt{Context in [CONTEXT\_LANG]: [CONTEXT]\textbackslash nNoisy text in [INPUT\_LANG]: [INPUT\_TEXT]\textbackslash nHow would you fix the [INPUT\_LANG] sentence to make the meaning the same as the context? [LABEL\_TEXT]} \\
            \midrule
            \texttt{[INPUT\_TEXT]. Denoise the previous [INPUT\_LANG] sentence to it equivalent sentence: [CONTEXT]\textbackslash n[LABEL\_TEXT]} \\
            \midrule
            \texttt{Context: [CONTEXT]\textbackslash nFix the following [INPUT\_LANG] text "[INPUT\_TEXT]" ensuring the meaning is equivalent with the context. [LABEL\_TEXT]} \\
            \midrule
            \texttt{Context: [CONTEXT]\textbackslash nNoisy text in [INPUT\_LANG]: [INPUT\_TEXT]\textbackslash nHow would you fix the [INPUT\_LANG] sentence to make the meaning the same as the [CONTEXT\_LANG] sentence? [LABEL\_TEXT]} \\
            \bottomrule
        \end{tabular}
    }
    \caption{Prompt used for Bilingual Denoising (\textbf{TLM}) task}
    \label{tab:tlm-prompt}
\end{table*}

\begin{table*}[!t]
    \centering
    \resizebox{0.9\linewidth}{!}{
        \begin{tabular}{p{\linewidth}}
            \toprule
            \textbf{Prompt in \colorbox{yellow!25}{Machine Translation (MT)} Task}  \\
            \midrule
            \texttt{Translate the following text from [SOURCE\_LANG] to [TARGET\_LANG].\textbackslash nText: [SOURCE\_TEXT]\textbackslash nTranslation: [TARGET\_TEXT]} \\
            \midrule
            \texttt{[SOURCE\_TEXT]\textbackslash nTranslate the text above from [SOURCE\_LANG] to [TARGET\_LANG]. [TARGET\_TEXT]} \\
            \midrule
            \texttt{Text in [SOURCE\_LANG]: [SOURCE\_TEXT]\textbackslash nHow would you translate that in [TARGET\_LANG]? [TARGET\_TEXT]} \\
            \midrule
            \texttt{Translate the following text to [TARGET\_LANG].\textbackslash nText: [SOURCE\_TEXT]\textbackslash nTranslation: [TARGET\_TEXT]} \\
            \midrule
            \texttt{[SOURCE\_TEXT]\textbackslash nTranslate the text above to [TARGET\_LANG]. [TARGET\_TEXT]} \\
            \midrule
            \texttt{Input text: [SOURCE\_TEXT]\textbackslash nHow would you translate that into [TARGET\_LANG]? [TARGET\_TEXT]} \\
            \bottomrule
        \end{tabular}
    }
    \caption{Prompt used for Machine Translation (\textbf{MT}) task}
    \label{tab:mt-prompt}
\end{table*}

\begin{table*}[!t]
    \centering
    \resizebox{0.9\linewidth}{!}{
        \begin{tabular}{p{\linewidth}}
            \toprule
            \textbf{Prompt in \colorbox{yellow!25}{Crosslingual Semantic Similarity (XSS)} Task}  \\
            \midrule
            \texttt{[SOURCE\_LANG] sentence: [SOURCE\_TEXT]\textbackslash n[TARGET\_LANG] sentence: [TARGET\_TEXT]\textbackslash nDo the two sentences have the same meaning? [LABEL]} \\
            \midrule
            \texttt{Sentence A: [SOURCE\_TEXT]\textbackslash nSentence B: [TARGET\_TEXT]\textbackslash nDo sentence A and sentence B have the same meaning? [LABEL]} \\
            \midrule
            \texttt{[SOURCE\_LANG] sentence: [SOURCE\_TEXT]\textbackslash n[TARGET\_LANG] sentence: [TARGET\_TEXT]\textbackslash nAre the two sentences equivalent? [LABEL]} \\
            \midrule
            \texttt{Sentence A: [SOURCE\_TEXT]\textbackslash nSentence B: [TARGET\_TEXT]\textbackslash nAre sentence A and sentence B equivalent? [LABEL]} \\
            \midrule
            \texttt{Is the [SOURCE\_LANG] sentence "[SOURCE\_TEXT]" equivalent to the [TARGET\_LANG] sentence "[TARGET\_TEXT]"? [LABEL]} \\
            \midrule
            \texttt{Is the sentence "[SOURCE\_TEXT]" equivalent to the sentence "[TARGET\_TEXT]"? [LABEL]} \\
            \bottomrule
        \end{tabular}
    }
    \caption{Prompt used for Crosslingual Semantic Similarity (\textbf{XSS}) task}
    \label{tab:xss-prompt}
\end{table*}

\begin{table*}[!t]
    \centering
    \resizebox{0.9\linewidth}{!}{
        \begin{tabular}{p{\linewidth}}
            \toprule
            \textbf{Prompt in \colorbox{yellow!25}{Monolingual Denoising (MLM)} Task}  \\
            \midrule
            \texttt{Denoise the following noisy [SOURCE\_LANG] text: "[SOURCE\_TEXT]",  to make a correct sentence. [TARGET\_TEXT]} \\
            \midrule
            \texttt{Fix and complete the following [SOURCE\_LANG] sentence: [SOURCE\_TEXT]\textbackslash n[TARGET\_TEXT]} \\
            \midrule
            \texttt{Sentence in [SOURCE\_LANG]: [SOURCE\_TEXT]\textbackslash nHow would you fix the sentence to make a correct sentence? [TARGET\_TEXT]} \\
            \midrule
            \texttt{Denoise the following noisy text "[SOURCE\_TEXT]" to make a correct [SOURCE\_LANG] sentence. [TARGET\_TEXT]} \\
            \midrule
            \texttt{Fix and complete the following sentence: [SOURCE\_TEXT]\textbackslash n[TARGET\_TEXT]} \\
            \midrule
            \texttt{Input text: [SOURCE\_TEXT]\textbackslash nHow would you fix the sentence to make a correct [SOURCE\_LANG] sentence? [TARGET\_TEXT]} \\
            \bottomrule
        \end{tabular}
    }
    \caption{Prompt used for Monolingual Denoising (\textbf{MLM}) task}
    \label{tab:mlm-prompt}
\end{table*}

\begin{table*}[!t]
    \centering
    \resizebox{0.95\linewidth}{!}{
        \begin{tabular}{p{\linewidth}}
            \toprule
            \textbf{Prompt in \colorbox{yellow!25}{Sentiment Analysis} Task}  \\
            \midrule
            \texttt{[INPUT]\textbackslash nWhat would be the sentiment of the text above? [OPTIONS]? [LABELS\_CHOICE]} \\
            \midrule
            \texttt{What is the sentiment of this text?\textbackslash nText: [INPUT]\textbackslash nAnswer with [OPTIONS]: [LABELS\_CHOICE]} \\
            \midrule
            \texttt{Text: [INPUT]\textbackslash n\textbackslash nPlease classify the sentiment of above text. Answer with [OPTIONS]: [LABELS\_CHOICE]} \\
            \bottomrule
        \end{tabular}
    }
    \caption{Prompt used for Sentiment Analysis task}
    \label{tab:senti-prompt}
\end{table*}

\begin{table*}[!t]
    \centering
    \resizebox{0.95\linewidth}{!}{
        \begin{tabular}{p{\linewidth}}
            \toprule
            \textbf{Prompt in \colorbox{yellow!25}{Emotion Recognition} Task}  \\
            \midrule
            \texttt{[INPUT]\textbackslash nWhat would be the emotion of the text above? [OPTIONS]? [LABELS\_CHOICE]} \\
            \midrule
            \texttt{What is the emotion of this text?\textbackslash nText: [INPUT]\textbackslash nAnswer with [OPTIONS]: [LABELS\_CHOICE]} \\
            \midrule
            \texttt{Text: [INPUT]\textbackslash n\textbackslash nPlease classify the emotion of above text. Answer with [OPTIONS]: [LABELS\_CHOICE]} \\
            \bottomrule
        \end{tabular}
    }
    \caption{Prompt used for Emotion Recognition task}
    \label{tab:emot-prompt}
\end{table*}

\begin{table*}[!t]
    \centering
    \resizebox{0.95\linewidth}{!}{
        \begin{tabular}{p{\linewidth}}
            \toprule
            \textbf{Prompt in \colorbox{yellow!25}{Topic Classification} Task}  \\
            \midrule
            \texttt{[INPUT]\textbackslash nWhat would be the topic of the text above? [OPTIONS]? [LABELS\_CHOICE]} \\
            \midrule
            \texttt{What is the topic of this text?\textbackslash nText: [INPUT]\textbackslash nAnswer with [OPTIONS]: [LABELS\_CHOICE]} \\
            \midrule
            \texttt{Text: [INPUT]\textbackslash n\textbackslash nPlease classify the topic of above text. Answer with [OPTIONS]: [LABELS\_CHOICE]} \\
            \bottomrule
        \end{tabular}
    }
    \caption{Prompt used for the Topic Classification task}
    \label{tab:topic-prompt}
\end{table*}

\section{Comparison Between LLM-int8() and Full Precision Inference}
\label{app:8bit-vs-32bit}

We run all inference within our experiment with 8-bit quantization using LLM.int8()~\cite{dettmers2022llmint8}. To the best of our knowledge, the effectiveness of LLM.int8()~\cite{dettmers2022llmint8} has never been evaluated on zero-shot prompting in low-resource language cases. We evaluate datasets from various Indonesian and local languages spoken in Indonesian which are listed in IndoNLU~\cite{wilie2020indonlu} and NusaCrowd~\cite{cahyawijaya2023nusacrowd}. Specifically, we evaluate on 10 languages in NusaX~\cite{winata2022nusax}, Javanese IMDB~\cite{wongso2021imdbjv}, IndoLEM Sentiment~\cite{koto2020indolem}, IndoNLI~\cite{mahendra2021indonli}, SmSA~\cite{purwarianti2019smsa}, CASA~\cite{ilmania2018absa}, and Sundanese Twitter Dataset for Emotion~\cite{putra2020suemot} datasets. Based on the result shown in Table~\ref{tab:precision-comparison}, there is only a marginal performance different between 8-bit quantization with LLM.int8() compared to the full precision models, which suggests the generalization of LLM.int8()~\cite{dettmers2022llmint8} for zero-shot prompting in low-resource languages.

\begin{table*}
    \centering
    \resizebox{0.95\linewidth}{!}{
        \begin{tabular}{lccccc}
            \toprule
            \textbf{Model} & \textbf{Prompt Lang.} & \textbf{Acc} & \textbf{Macro F1} & \textbf{Macro Prec.} & \textbf{Macro Rec.} \\
            \midrule 
            \multicolumn{6}{c}{\textit{Full Precision}} \\
            \midrule 
            BLOOMZ-560M & EN & 47.58 & 33.25 & 37.97 & 43.11 \\
            BLOOMZ-560M & ID & 44.37 & 29.78 & 37.79 & 40.28 \\
            BLOOMZ-1B1 & EN & 52.26 & 37.90 & 40.48 & 45.79 \\
            BLOOMZ-1B1 & ID & 52.88 & 39.28 & 46.42 & 46.67 \\
            BLOOMZ-1B7 & EN & 51.44 & 36.90 & 41.90 & 45.10 \\
            BLOOMZ-1B7 & ID & 52.68 & 41.20 & 50.81 & 48.03 \\
            \midrule 
            \multicolumn{6}{c}{\textit{8-Bit Quantization}} \\
            \midrule 
            BLOOMZ-560M & EN & 47.56 & 34.67 & 40.94 & 42.97 \\
            BLOOMZ-560M & ID & 43.64 & 33.30 & 42.90 & 39.68 \\
            BLOOMZ-1B1 & EN & 50.68 & 37.52 & 40.37 & 44.56 \\
            BLOOMZ-1B1 & ID & 51.23 & 38.69 & 43.53 & 45.34 \\
            BLOOMZ-1B7 & EN & 49.71 & 35.05 & 42.11 & 43.57 \\
            BLOOMZ-1B7 & ID & 52.61 & 41.87 & 51.74 & 48.15 \\
            BLOOMZ-3B & EN & 54.80 & 40.78 & 46.59 & 48.24 \\
            BLOOMZ-3B & ID & 56.75 & 44.34 & 45.16 & 51.12 \\
            \bottomrule
        \end{tabular}
    }
    \caption{Evaluation of full precision and 8-bit quantization on various Indonesian local languages datasets.}
    \label{tab:precision-comparison}
\end{table*}

\section{Datasets Details}
\label{app:datasets}

In this section, we describe the statistics for each dataset use in the experiment. Table~\ref{tab:nusamenulis-dataset} shows the statistics for the sentiment analysis task of NusaTranslation~\cite{cahyawijaya2023nusawrites}. For the Indonesian subset, we take the first fold of the IndoLEM sentiment~\cite{koto2020indolem}, which is the Indonesian sentiment analysis dataset used as the source sentences in the NusaTranslation~\cite{cahyawijaya2023nusawrites}. Table~\ref{tab:nusax-dataset} shows the statistics for the sentiment analysis task of NusaX~\cite{winata2022nusax}. Table~\ref{tab:nusa-para-er-dataset} and Table~\ref{tab:nusa-para-tc-dataset} display the statistics for the emotion recognition and topic classification tasks of NusaParagraph~\cite{cahyawijaya2023nusawrites}, respectively. 

\begin{table}[!t]
    \centering
    \resizebox{\linewidth}{!}{
    \begin{tabular}{l|c|c|c|c}
    \toprule
    \textbf{Status} &\textbf{Language} & \textbf{Train} &\textbf{Valid.} &\textbf{Test} \\ 
    \midrule
    \multirow{1}{*}{Pre-trained}  &Indonesian (ind) & 3638 & 399 & 1011 \\
    \midrule
    \multirow{3}{*}{Seen}   &Javanese (jav)     & 3400  & 448 & 1200 \\
                            &Sundanese (sun)    & 3400  & 448 & 1200 \\
                            &Minangkabau (min)  & 3400  & 448 & 1200 \\
    \midrule
    \multirow{7}{*}{Unseen} &Ambon (abs)        & 250   & 98  & 500  \\
                            &Batak (btk)        & 3400  & 448 & 1200 \\
                            &Betawi (bew)       & 3400  & 448 & 1200 \\
                            &Bima (bhp)         & 260   & 100 & 500  \\
                            &Madurese (mad)     & 3400  & 448 & 1200 \\
                            &Makassarese (mak)  & 3400  & 448 & 1200 \\
                            &Musi (mui)         & 250	& 91  & 500  \\
                            &Rejang (rej)       & 250	& 78  & 500  \\
    \bottomrule
    \end{tabular}
    }
    \caption{Statistics of NusaTranslation sentiment analysis dataset. \textbf{Pre-trained} denotes languages that are already seen before the InstructAlign tuning. \textbf{Seen} denotes languages that are seen during the InstructAlign.\textbf{Unseen} denotes languages that are still unseen after the InstructAlign.}
    \label{tab:nusamenulis-dataset}
\end{table}

\begin{table}[!t]
    \centering
    \resizebox{\linewidth}{!}{
    \begin{tabular}{l|c|c|c|c}
    \toprule
    \textbf{Status} &\textbf{Language} & \textbf{Train} &\textbf{Valid.} &\textbf{Test} \\ 
    \midrule
    \multirow{2}{*}{Pre-trained}  &English (eng)        &500	&100  &400 \\
                                  &Indonesia (ind)      &500	  &100  &400 \\
    \midrule
    \multirow{8}{*}{Seen}   &Aceh (ace)     &500	&100  &400 \\
                            &Bali (ban)     &500	&100  &400 \\
                            &Banjar (bjn)   &500	&100  &400 \\
                            &Bugis (bug)    &500	&100  &400 \\
                            &Minang (min)   &500	&100  &400 \\
                            &Javanese(jav)  &500	&100  &400 \\
                            &Sunda (sun)    &500	&100  &400 \\
    \midrule
    \multirow{3}{*}{Unseen} &Madura (mad)   &500	&100  &400 \\
                            &Ngaju (nij)    &500	&100  &400 \\
                            &Bataknese (bbc)&500	&100  &400 \\
    \bottomrule
    \end{tabular}
    }
    \caption{Statistics of NusaX sentiment analysis dataset. \textbf{Pre-trained} denotes languages that are already seen before the InstructAlign. \textbf{Seen} denotes languages that are seen during the InstructAlign.\textbf{Unseen} denotes languages that are still unseen after the InstructAlign.}
    \label{tab:nusax-dataset}
\end{table}

\begin{table}[!t]
    \centering
    \resizebox{\linewidth}{!}{
    \begin{tabular}{l|c|c|c|c}
    \toprule
    \textbf{Status} &\textbf{Language} & \textbf{Train} &\textbf{Valid.} &\textbf{Test} \\ 
    \midrule
    \multirow{4}{*}{Unseen}&Javanese (jav)      &2800	&440  &800 \\
                           &Minangkabau (min)   &2000	 &357  &800 \\
                           &Sundanese  (sun)    &2400	 &400  &800 \\
                           &Buginese	(bug)    &87	 &50   &300 \\
    \midrule
    \multirow{6}{*}{Seen}  &Batak	   (btk)    &1150   &292  &500 \\
                           &Betawi	    (bew)    &2700	 &430  &800 \\
                           &Madurese	(mad)    &1000	 &263  &500 \\
                           &Makassarese(mak)    &1500	 &304  &500 \\
                           &Musi       (mui)    &200	 &75   &400 \\
                           &Rejang     (rej)    &136	 &50   &300 \\
    \bottomrule
    \end{tabular}
    }
    \caption{Statistics of NusaParagraph emotion recognition dataset. \textbf{Pre-trained} denotes languages that are already seen before InstructAlign. \textbf{Seen} denotes languages that are seen during InstructAlign.\textbf{Unseen} denotes languages that are still unseen after InstructAlign.}
    \label{tab:nusa-para-er-dataset}
\end{table}

\begin{table}[!t]
    \centering
    \resizebox{\linewidth}{!}{
    \begin{tabular}{l|c|c|c|c}
    \toprule
    \textbf{Status} &\textbf{Language} & \textbf{Train} &\textbf{Valid.} &\textbf{Test} \\ 
    \midrule
    \multirow{4}{*}{Unseen}&Javanese (jav)      &2650	&448  &800 \\
                           &Minangkabau (min)   &2400	 &399  &800 \\
                           &Sundanese  (sun)    &2800	 &468  &900 \\
                           &Buginese	(bug)    &93	 &50   &300 \\
    \midrule
    \multirow{6}{*}{Seen}  &Batak	   (btk)    &1350   &275  &500 \\
                           &Betawi	    (bew)    &2650	 &435  &800 \\
                           &Madurese	(mad)    &1800	 &367  &700 \\
                           &Makassarese(mak)    &1500	 &376  &700 \\
                           &Musi       (mui)    &168	 &80   &400 \\
                           &Rejang     (rej)    &105	 &50   &350 \\
    \bottomrule
    \end{tabular}
    }
    \caption{Statistics of NusaParagraph topic classification dataset. \textbf{Pre-trained} denotes languages that are already seen before InstructAlign. \textbf{Seen} denotes languages that are seen during InstructAlign.\textbf{Unseen} denotes languages that are still unseen after InstructAlign.}
    \label{tab:nusa-para-tc-dataset}
\end{table}

\section{Detailed Experiment Results}
\label{app:results}

In this section, we provide the complete experimental result per dataset.  Table~\ref{tab:nusatranslation-result} shows the experiment results on the sentiment analysis task of NusaTranslation. Table~\ref{tab:nusax-result} shows the experiment results on the sentiment analysis task of NusaX~\cite{winata2022nusax}. Table~\ref{tab:nusa-para-er-result} and Table~\ref{tab:nusa-para-tc-result} show the experiment results on the emotion recognition and topic classification tasks of NusaParagraph, respectively. 

\begin{table*}[!t]
    \centering
    \resizebox{\linewidth}{!}{
        \begin{tabular}{l|c|c|c|c|c|c|c|c|c|c|c|c}
        \toprule
        \multicolumn{1}{c|}{\multirow{2}{*}{\textbf{Model}}} & \textbf{L1} & \multicolumn{3}{c|}{\textbf{L2}} & \multicolumn{8}{c}{\textbf{L3}} \\
        \cmidrule{2-13}
         & \textbf{ind} & \textbf{jav} & \textbf{min} & \textbf{sun} & \textbf{abs} & \textbf{bew} & \textbf{bhp} & \textbf{btk} & \textbf{mad} & \textbf{mak} & \textbf{mui} & \textbf{rej} \\
         \midrule
        BLOOM 560m & 61.47 & 56.09 & 58.13 & 58.63 & 62.53 & 58.42 & 49.72 & 56.05 & 54.02 & 53.97 & 60.27 & 55.55 \\
        BLOOM 1b1 & 58.81 & 59.05 & 59.33 & 59.16 & 47.87 & 58.23 & 60.85 & 58.95 & 58.79 & 58.83 & 54.92 & 57.73 \\
        BLOOM 3b & 58.30 & 44.84 & 45.48 & 44.61 & 46.08 & 45.61 & 44.54 & 43.62 & 43.36 & 44.03 & 45.05 & 43.15 \\
        \midrule         
        BLOOMZ 560m & 69.81 & 43.00 & 50.97 & 46.51 & 45.23 & 47.87 & 33.13 & 36.69 & 36.84 & 35.30 & 61.42 & 36.21 \\
        BLOOMZ 1b1 & 80.40 & 61.32 & 68.95 & 61.75 & 61.07 & 66.94 & 46.18 & 50.20 & 49.71 & 50.66 & 70.31 & 52.01 \\
        BLOOMZ 3b & 81.38 & 68.05 & 71.76 & 68.43 & 69.57 & 69.76 & 67.73 & 65.09 & 64.37 & 63.14 & 69.08 & 64.05 \\
        \midrule
        MLM BLOOMZ 560m & 65.68 & 23.29 & 21.11 & 22.31 & 20.86 & 22.00 & 20.40 & 19.04 & 21.10 & 20.59 & 28.82 & 19.50 \\
        MLM BLOOMZ 560m-r=100k & 71.93 & 63.89 & 69.37 & 66.27 & 64.46 & 65.38 & 56.12 & 62.68 & 58.20 & 56.51 & 67.73 & 58.12 \\
        MLM BLOOMZ 1b1-r=100k & 73.25 & 71.18 & 72.24 & 70.95 & 62.67 & 67.65 & 56.07 & 59.22 & 58.60 & 60.38 & 68.10 & 60.27 \\
        \midrule
        MT BLOOMZ 560m & 55.20 & 41.87 & 39.00 & 38.16 & 36.88 & 39.29 & 36.07 & 34.74 & 36.97 & 33.81 & 41.72 & 37.70 \\
        MT BLOOMZ 560m-r=100k & 74.46 & 70.73 & 69.94 & 70.00 & 66.81 & 67.65 & 64.58 & 66.53 & 65.10 & 61.35 & 68.43 & 63.23 \\
        MT BLOOMZ 1b1-r=100k & 70.86 & 59.62 & 62.22 & 61.63 & 54.37 & 57.97 & 49.95 & 50.22 & 51.31 & 52.22 & 60.25 & 50.30 \\
        \midrule
        TLM BLOOMZ 560m & 71.57 & 66.74 & 66.05 & 66.94 & 63.06 & 65.64 & 59.00 & 61.07 & 61.31 & 61.13 & 65.30 & 63.17 \\
        TLM BLOOMZ 560m-r=1k & 70.52 & 61.73 & 62.76 & 62.01 & 54.34 & 56.31 & 48.52 & 49.44 & 49.21 & 47.95 & 61.11 & 49.22 \\
        TLM BLOOMZ 560m-r=10k & 72.82 & 66.27 & 66.22 & 66.76 & 62.29 & 63.32 & 59.61 & 61.27 & 60.35 & 60.32 & 63.60 & 60.49 \\
        TLM BLOOMZ 560m-r=100k & 72.40 & 61.05 & 59.43 & 62.11 & 54.51 & 56.44 & 46.68 & 50.72 & 50.56 & 45.02 & 63.27 & 48.39 \\
        TLM BLOOMZ 1b1-r=100k & 75.66 & 70.05 & 70.12 & 70.70 & 64.47 & 67.07 & 62.92 & 61.87 & 60.96 & 61.86 & 68.11 & 61.53 \\
        \midrule
        XSS BLOOMZ 560m & 64.48 & 57.65 & 52.18 & 54.13 & 52.40 & 53.59 & 48.55 & 48.06 & 49.59 & 44.01 & 58.03 & 49.43 \\
        XSS BLOOMZ 560m-r=1k & 69.34 & 63.55 & 62.84 & 65.45 & 65.20 & 64.15 & 59.11 & 60.53 & 62.34 & 61.58 & 63.51 & 58.36 \\
        XSS BLOOMZ 560m-r=10k & 72.22 & 67.89 & 67.81 & 67.25 & 62.76 & 64.42 & 62.83 & 61.98 & 62.02 & 62.21 & 65.38 & 59.27 \\
        XSS BLOOMZ 560m-r=100k & 71.27 & 68.34 & 67.89 & 68.07 & 61.58 & 68.69 & 62.66 & 65.73 & 63.44 & 58.24 & 70.24 & 64.92 \\
        XSS BLOOMZ 1b1-r=100k & 76.75 & 72.40 & 71.40 & 71.87 & 63.75 & 65.45 & 60.27 & 61.27 & 60.10 & 63.21 & 66.16 & 60.49 \\
        \bottomrule
        \end{tabular}
    }
    \caption{Experiment result on the sentiment analysis task of the NusaTranslation dataset}
    \label{tab:nusatranslation-result}
\end{table*}

\begin{table*}[!t]
    \centering
    \resizebox{\linewidth}{!}{
        \begin{tabular}{l|c|c|c|c|c|c|c|c|c|c|c|c}
        \toprule
        \multicolumn{1}{c|}{\multirow{2}{*}{\textbf{Model}}} & \multicolumn{2}{c|}{\textbf{L1}} & \multicolumn{7}{c|}{\textbf{L2}} & \multicolumn{3}{c}{\textbf{L3}} \\
        \cmidrule{2-13}
         & \textbf{eng} & \textbf{ind} & \textbf{ace} & \textbf{ban} & \textbf{bjn} & \textbf{bug} & \textbf{jav} & \textbf{min} & \textbf{sun} & \textbf{bbc} & \textbf{mad} & \textbf{nij} \\
         \midrule
        BLOOM 560m & 29.26 & 21.13 & 21.35 & 21.93 & 21.35 & 23.21 & 21.86 & 21.82 & 21.04 & 22.28 & 22.13 & 21.11 \\
        BLOOM 1b1 & 22.02 & 22.54 & 21.47 & 22.62 & 22.27 & 21.34 & 22.97 & 21.92 & 21.55 & 22.10 & 21.65 & 21.53 \\
        BLOOM 3b & 24.03 & 21.17 & 21.31 & 21.17 & 21.18 & 21.35 & 21.17 & 21.17 & 21.17 & 21.19 & 21.20 & 21.17 \\
        \midrule
        BLOOMZ 560m & 58.24 & 55.59 & 31.18 & 32.40 & 37.17 & 27.79 & 35.86 & 39.29 & 32.44 & 29.49 & 32.80 & 38.15 \\
        BLOOMZ 1b1 & 57.41 & 58.58 & 43.31 & 43.02 & 44.72 & 31.12 & 46.52 & 42.59 & 39.20 & 26.82 & 41.92 & 40.76 \\
        BLOOMZ 3b & 62.65 & 63.21 & 48.81 & 48.40 & 55.27 & 23.47 & 54.26 & 51.11 & 39.41 & 32.42 & 38.88 & 41.68 \\
        \midrule
        MLM BLOOMZ 560m & 49.99 & 49.33 & 31.74 & 28.37 & 34.32 & 25.76 & 33.89 & 31.27 & 29.20 & 28.43 & 32.08 & 30.98 \\
        MLM BLOOMZ 560m-R-100000 & 61.32 & 60.01 & 42.69 & 41.69 & 50.95 & 31.53 & 44.28 & 44.30 & 42.11 & 33.18 & 41.05 & 40.15 \\
        MLM BLOOMZ 1b1-R-100000 & 61.30 & 59.73 & 43.11 & 43.02 & 50.71 & 31.31 & 53.66 & 51.05 & 47.27 & 31.13 & 42.02 & 39.83 \\
        \midrule
        MT BLOOMZ 560m & 47.24 & 41.41 & 31.78 & 33.78 & 34.69 & 28.44 & 35.47 & 35.15 & 36.01 & 26.86 & 26.69 & 27.49 \\
        MT BLOOMZ 560m-R-100000 & 60.09 & 54.18 & 39.11 & 42.59 & 46.22 & 34.50 & 43.37 & 41.31 & 41.31 & 35.95 & 38.54 & 39.84 \\
        MT BLOOMZ 1b1-R-100000 & 59.18 & 53.69 & 43.97 & 45.40 & 50.16 & 38.65 & 48.37 & 45.97 & 41.98 & 37.97 & 40.90 & 40.60 \\
        \midrule
        TLM BLOOMZ 560m & 44.72 & 46.02 & 33.59 & 34.26 & 41.16 & 25.36 & 41.76 & 38.72 & 37.40 & 25.67 & 30.88 & 29.98 \\
        TLM BLOOMZ 560m-R-1000 & 58.05 & 54.59 & 43.03 & 37.06 & 46.55 & 34.02 & 43.21 & 43.24 & 39.59 & 33.99 & 38.16 & 37.39 \\
        TLM BLOOMZ 560m-R-10000 & 57.38 & 57.73 & 43.43 & 36.76 & 45.99 & 35.06 & 44.38 & 43.30 & 40.83 & 34.06 & 42.46 & 40.00 \\
        TLM BLOOMZ 560m-R-100000 & 61.65 & 56.50 & 41.78 & 41.36 & 48.15 & 31.19 & 48.89 & 44.12 & 44.90 & 33.78 & 41.51 & 37.90 \\
        TLM BLOOMZ 1b1-R-100000 & 64.26 & 63.54 & 52.22 & 51.35 & 58.19 & 41.87 & 59.48 & 59.67 & 56.99 & 38.26 & 48.11 & 48.01 \\
        \midrule
        XSS BLOOMZ 560m & 53.93 & 53.19 & 43.60 & 41.73 & 47.09 & 37.79 & 47.29 & 45.36 & 43.42 & 32.59 & 41.66 & 40.79 \\
        XSS BLOOMZ 560m-R-1000 & 56.57 & 54.90 & 36.78 & 40.28 & 42.20 & 28.56 & 45.67 & 41.33 & 39.80 & 27.30 & 31.67 & 32.20 \\
        XSS BLOOMZ 560m-R-10000 & 55.62 & 57.84 & 44.24 & 44.03 & 50.04 & 32.87 & 48.92 & 45.55 & 45.64 & 36.38 & 40.36 & 43.12 \\
        XSS BLOOMZ 560m-R-100000 & 59.89 & 58.22 & 45.53 & 39.57 & 52.68 & 36.15 & 49.83 & 50.61 & 46.45 & 35.27 & 42.40 & 43.39 \\
        XSS BLOOMZ 1b1-R-100000 &  60.78 & 59.34 & 45.83 & 45.45 & 53.08 & 36.24 & 52.24 & 50.54 & 47.20 & 33.81 & 40.99 & 41.08 \\
        \bottomrule
        \end{tabular}
    }
    \caption{Experiment result on the sentiment analysis task of the NusaX dataset}
    \label{tab:nusax-result}
\end{table*}

\begin{table*}[!t]
    \centering
    \resizebox{\linewidth}{!}{
        \begin{tabular}{l|c|c|c|c|c|c|c|c|c|c}
        \toprule
        \multicolumn{1}{c|}{\multirow{2}{*}{\textbf{Model}}} & \multicolumn{4}{c|}{\textbf{L2}} & \multicolumn{6}{c}{\textbf{L3}} \\
        \cmidrule{2-11}
         & \textbf{bug} & \textbf{jav} & \textbf{min} & \textbf{sun} & \textbf{bew} & \textbf{btk} & \textbf{mad} & \textbf{mak} & \textbf{mui} & \textbf{rej}\\
         \midrule
        BLOOM 560m & 1.19 & 2.42 & 4.54 & 3.05 & 4.37 & 2.56 & 0.59 & 1.42 & 1.11 & 2.66 \\
        BLOOM 1b1 & 1.19 & 2.42 & 4.54 & 3.05 & 4.29 & 2.57 & 0.59 & 1.42 & 1.11 & 2.44 \\
        BLOOM 3b & 1.19 & 2.42 & 4.54 & 3.05 & 4.29 & 2.57 & 0.59 & 1.42 & 1.11 & 2.44 \\
        \midrule
        BLOOMZ 560m & 2.36 & 2.93 & 4.71 & 3.52 & 4.35 & 3.33 & 1.41 & 3.09 & 1.28 & 4.10 \\
        BLOOMZ 1b1 & 1.19 & 2.42 & 4.54 & 3.05 & 4.29 & 2.57 & 0.59 & 1.42 & 1.11 & 2.44 \\
        BLOOMZ 3b & 1.19 & 2.42 & 4.54 & 3.05 & 4.29 & 2.57 & 0.59 & 1.42 & 1.11 & 2.44 \\
        \midrule
        MLM BLOOMZ 560m & 1.19 & 2.51 & 4.63 & 3.04 & 4.29 & 2.57 & 0.59 & 1.42 & 1.11 & 2.44 \\
        MLM BLOOMZ 560m-R-100000 & 1.19 & 2.41 & 4.54 & 3.05 & 4.29 & 2.57 & 0.59 & 1.42 & 1.11 & 2.44 \\
        MLM BLOOMZ 1b1-R-100000 & 1.19 & 2.42 & 4.71 & 3.05 & 4.29 & 2.57 & 0.59 & 1.42 & 1.11 & 2.44 \\
        \midrule
        MT BLOOMZ 1b1-R-100000 & 1.60 & 2.76 & 4.77 & 3.54 & 4.27 & 2.56 & 0.59 & 1.42 & 1.12 & 2.44 \\
        MT BLOOMZ 560m & 1.41 & 2.58 & 9.14 & 5.63 & 4.45 & 2.57 & 0.59 & 1.56 & 2.10 & 2.44 \\
        MT BLOOMZ 560m-R-100000 & 1.19 & 2.51 & 4.54 & 3.04 & 4.29 & 2.70 & 0.59 & 1.42 & 1.11 & 2.44 \\
        \midrule
        TLM BLOOMZ 560m & 1.19 & 2.58 & 4.88 & 3.54 & 4.29 & 2.57 & 0.59 & 1.42 & 1.11 & 2.44 \\
        TLM BLOOMZ 560m-R-1000 & 1.19 & 2.50 & 5.10 & 3.14 & 4.29 & 2.57 & 0.59 & 1.42 & 1.12 & 2.44 \\
        TLM BLOOMZ 560m-R-10000 & 1.19 & 2.67 & 5.12 & 4.34 & 4.29 & 2.57 & 0.73 & 1.42 & 1.11 & 2.66 \\
        TLM BLOOMZ 560m-R-100000 & 1.40 & 2.42 & 4.54 & 3.13 & 4.29 & 2.71 & 0.73 & 1.42 & 1.11 & 2.44 \\
        TLM BLOOMZ 1b1-R-100000 & 1.19 & 2.41 & 4.63 & 3.13 & 4.29 & 2.57 & 0.59 & 1.42 & 1.12 & 2.44 \\
        \midrule
        XSS BLOOMZ 560m & 1.54 & 3.12 & 4.65 & 3.77 & 4.29 & 2.71 & 0.59 & 1.42 & 1.11 & 2.44 \\
        XSS BLOOMZ 560m-R-1000 & 1.56 & 2.82 & 5.16 & 3.54 & 4.37 & 2.69 & 0.73 & 1.42 & 1.11 & 2.44 \\
        XSS BLOOMZ 560m-R-10000 & 1.39 & 2.84 & 5.56 & 4.10 & 4.29 & 2.57 & 0.59 & 1.43 & 1.28 & 2.44 \\
        XSS BLOOMZ 560m-R-100000 & 1.19 & 2.42 & 4.54 & 3.05 & 4.29 & 2.57 & 0.59 & 1.42 & 1.11 & 2.42 \\
        XSS BLOOMZ 1b1-R-100000 & 1.19 & 2.67 & 4.63 & 3.84 & 4.29 & 2.57 & 0.59 & 1.42 & 1.11 & 2.44 \\
        \bottomrule
        \end{tabular}
    }
    \caption{Experiment result on the emotion recognition task of the NusaParagraph dataset}
    \label{tab:nusa-para-er-result}
\end{table*}

\begin{table*}[!t]
    \centering
    \resizebox{\linewidth}{!}{
        \begin{tabular}{l|c|c|c|c|c|c|c|c|c|c}
        \toprule
        \multicolumn{1}{c|}{\multirow{2}{*}{\textbf{Model}}} & \multicolumn{4}{c|}{\textbf{L2}} & \multicolumn{6}{c}{\textbf{L3}} \\
        \cmidrule{2-11}
         & \textbf{bug} & \textbf{jav} & \textbf{min} & \textbf{sun} & \textbf{bew} & \textbf{btk} & \textbf{mad} & \textbf{mak} & \textbf{mui} & \textbf{rej}\\
         \midrule
        BLOOM-560m & 7.68 & 3.50 & 6.36 & 3.80 & 5.42 & 7.92 & 11.25 & 9.07 & 3.91 & 5.80 \\
        BLOOM-1b1 & 7.72 & 3.50 & 6.36 & 3.81 & 5.42 & 7.93 & 11.26 & 9.09 & 3.91 & 5.82 \\
        BLOOM-3b & 7.72 & 3.50 & 6.36 & 3.81 & 5.42 & 7.93 & 11.26 & 9.09 & 3.91 & 5.82 \\
        \midrule
        BLOOMZ-560m & 9.13 & 4.10 & 6.86 & 4.29 & 6.07 & 8.75 & 11.71 & 9.45 & 4.09 & 6.04 \\
        BLOOMZ-1b1 & 7.72 & 3.50 & 6.36 & 3.81 & 5.51 & 7.93 & 11.26 & 9.09 & 3.91 & 5.82 \\
        BLOOMZ-3b & 7.72 & 4.21 & 6.70 & 4.30 & 7.55 & 8.33 & 11.28 & 9.19 & 7.30 & 5.82 \\
        \midrule
        MLM BLOOMZ-560m & 8.15 & 3.50 & 6.36 & 3.81 & 5.42 & 7.93 & 11.26 & 9.09 & 3.91 & 5.82 \\
        MLM BLOOMZ-560m $r$=100000 & 7.72 & 3.49 & 6.52 & 4.36 & 5.51 & 7.93 & 11.27 & 9.09 & 4.08 & 5.82 \\
        MLM BLOOMZ-1b1 $r$=100000 & 7.72 & 3.50 & 6.36 & 3.81 & 5.42 & 7.93 & 11.26 & 9.09 & 3.91 & 5.82 \\
        \midrule
        MT BLOOMZ-560m & 7.72 & 3.50 & 6.36 & 3.81 & 5.42 & 7.93 & 11.26 & 9.09 & 3.92 & 5.82 \\
        MT BLOOMZ-560m $r$=100000 & 7.72 & 3.58 & 6.37 & 3.93 & 5.52 & 7.94 & 11.22 & 9.19 & 3.92 & 5.82 \\
        MT BLOOMZ-1b1 $r$=100000 & 8.61 & 4.59 & 7.08 & 5.08 & 5.71 & 8.20 & 11.52 & 9.29 & 4.63 & 6.01 \\
        \midrule
        TLM BLOOMZ-560m & 9.43 & 3.83 & 7.27 & 7.11 & 5.42 & 7.93 & 11.28 & 9.18 & 3.92 & 5.82 \\
        TLM BLOOMZ-560m $r$=1000 & 14.08 & 11.46 & 17.31 & 16.55 & 10.35 & 12.61 & 11.92 & 12.04 & 9.34 & 5.96 \\
        TLM BLOOMZ-560m $r$=10000 & 8.37 & 4.23 & 7.66 & 5.40 & 5.43 & 8.05 & 11.20 & 9.28 & 4.25 & 5.96 \\
        TLM BLOOMZ-560m $r$=100000 & 7.75 & 3.50 & 6.34 & 3.80 & 5.51 & 7.93 & 11.35 & 9.18 & 4.23 & 5.78 \\
        TLM BLOOMZ-1b1 $r$=100000 & 7.71 & 3.67 & 6.55 & 4.05 & 5.42 & 7.93 & 11.27 & 9.08 & 3.91 & 5.82 \\
        \midrule
        XSS BLOOMZ-560m & 8.38 & 3.57 & 6.46 & 3.88 & 5.42 & 7.94 & 11.26 & 9.09 & 3.91 & 5.82 \\
        XSS BLOOMZ-560m $r$=1000 & 6.14 & 4.21 & 4.34 & 6.14 & 4.38 & 7.29 & 11.21 & 8.39 & 5.52 & 6.63 \\
        XSS BLOOMZ-560m $r$=10000 & 8.06 & 4.24 & 7.46 & 5.07 & 5.41 & 8.23 & 11.32 & 9.10 & 4.08 & 5.85 \\
        XSS BLOOMZ-560m $r$=100000 & 7.73 & 3.50 & 6.67 & 4.23 & 5.50 & 7.93 & 11.26 & 9.09 & 3.92 & 5.82 \\
        XSS BLOOMZ-1b1 $r$=100000 & 8.00 & 4.05 & 7.40 & 4.62 & 5.67 & 8.08 & 11.55 & 9.19 & 4.08 & 5.83 \\
        \bottomrule
        \end{tabular}
    }
    \caption{Experiment result on the topic classification task of the NusaParagraph dataset}
    \label{tab:nusa-para-tc-result}
\end{table*}

\end{document}